\documentclass[11pt,a4paper]{article}
\usepackage[hyperref]{emnlp2020}
\usepackage{times}
\usepackage{latexsym}
\usepackage{amsmath}

\usepackage{amsfonts}
\usepackage{graphicx}
\usepackage{subfigure,booktabs}
\usepackage{url}
\usepackage{multirow}
\usepackage{soul}
\usepackage[norelsize]{algorithm2e}

\usepackage{microtype}

\aclfinalcopy % Uncomment this line for the final submission
\title{Universal Natural Language Processing with Limited Annotations:\\Try Few-shot Textual Entailment as a Start}

\author{Wenpeng Yin$^1$, Nazneen Fatema Rajani$^1$, \\
\textbf{Dragomir Radev$^{1,2}$}, \textbf{Richard Socher}$^1$ and \textbf{Caiming Xiong$^1$} \\
  $^1$Salesforce Research and $^2$Yale University\\
  {\tt
    wyin@salesforce.com}}
    
\date{}

\newcommand{\modelname}{\textsc{UFO-Entail}}
\newcommand{\roberta}{RoBERTa}
\begin{document}
\maketitle

\begin{abstract}
A standard way to address different NLP problems is by first constructing a problem-specific dataset, then building a model to fit this dataset. To build the ultimate artificial intelligence, we desire a single machine that can handle diverse new problems, for which task-specific annotations are limited. We bring up textual entailment as a unified solver for such NLP problems. 
However, current research of textual entailment has not spilled much ink on the following questions: (i) How well does a pretrained textual entailment system generalize across domains with only a handful of domain-specific examples? and  (ii) When is it worth transforming an NLP task into textual entailment? We argue that the transforming is unnecessary if we can obtain rich annotations for this task. Textual entailment really matters particularly when the target NLP task has insufficient annotations. 

Universal NLP\footnote{``Universal NLP'' here means using a single machine  to address diverse NLP problems. This is different from using the same machine learning algorithm such as convolution nets to solve tasks because the latter still results in task-specific models which can not solve other tasks.} can be probably achieved through different routines. In this work, we introduce  \textbf{U}niversal \textbf{F}ew-sh\textbf{o}t textual \textbf{Entail}ment (\modelname). We demonstrate that this framework enables a pretrained entailment model to work well
on  new entailment domains in a few-shot setting, and show its effectiveness  as a unified solver for several downstream NLP tasks such as question answering and  coreference resolution  when the end-task annotations are limited. Code: \url{https://github.com/salesforce/UniversalFewShotNLP}
\end{abstract}

\section{Introduction}

Nowadays, the whole NLP journey has been broken down into innumerable sub-tasks. We often solve each task separately by first gathering task-specific training data and then tuning a machine learning system to learn the patterns in the data. Constrained by the  current techniques, the journey has to be performed  in this way. By a forward-looking perspective, instead, a single machine that can handle diverse (seen and unseen) tasks is desired. The reason is that we cannot always rely on expensive human resources to annotate large-scale task-specific labeled data, especially  considering the inestimable number of tasks to be explored. Therefore, a reasonable attempt is to map diverse NLP tasks into a common learning problem---solving this common problem equals to solving any downstream NLP tasks, even some tasks that are new or have insufficient annotations.

Textual entailment (aka. natural language inference in \citet{DBLPBowmanAPM15}) is the task of studying the relation of two assertive sentences, Premise (P) and Hypothesis (H): whether H is true given P.   Textual entailment (TE) was originally brought up as a unified framework for modeling diverse NLP tasks \citep{DBLPDaganGM05,DBLPPoliakHRHPWD18}. The research on TE dates back more than two decades and has made significant progress. Particularly, with the advances of deep neural networks and the availability of large-scale human annotated datasets,  fine-tuned systems often claim surpassing human performance on certain benchmarks.
% \footnote{Refer to the GLUE leaderboard: \url{https://gluebenchmark.com/leaderboard/}}

Nevertheless, two open problems remain. First, the increasing performances on some benchmarks heavily rely on rich human annotations. There is rarely a trained entailment system that can work on benchmarks in other domains. Current  textual entailment systems are far from being deployed in  new domains where no rich annotation exists.  Second, there is an increasing awareness in the community that lots of NLP tasks can be studied in the entailment framework. But it is unclear when it is worth transforming a target NLP tasks to textual entailment. We argue that textual entailment particularly  matters  when the target NLP task has insufficient annotations; in this way, some NLP tasks that share the same inference pattern and annotations are insufficient to build a task-specific model can be handled by a unified entailment system.

% there is an increasingly unclear connection between textual entailment and downstream NLP challenges since more and more NLI datasets are created in isolation of any end NLP tasks. We need to always keep in mind that the study of textual entailment is for universal NLP; this is even more crucial when we aim to tackle real world challenges where  annotations are scarce.

Motivated by the two issues,  we build \modelname---the first ever generalized few-shot textual entailment system with the following setting. We first assume that we can access a large-scale generic purpose TE dataset, such as MNLI \citep{DBLPWilliamsNB18}; this dataset enables us to build a  base entailment system with acceptable performance. To get even better performance in any new domain or new task, we combine the  generic purpose TE dataset with a couple of domain/task-specific examples to learn a better-performing entailment for that new domain/task. This is a reasonable assumption because in the real-world, any new domain or new task does not typically have large annotated data, but obtaining a couple of examples is usually feasible.

Technically, our \modelname\enspace is inspired by   the   Prototypical Network \cite{DBLPSnellSZ17}, a popular metric-based meta-learning paradigm, and the STILTS \cite{DBLP01088}, a framework that makes use of pretraining on indirect tasks to help the target task. \modelname~ consists of a RoBERTa \cite{DBLP11692} encoder and a proposed cross-task nearest neighbor block. The RoBERTa, pretrained on MNLI, provides a representation space biased to the source domain; the cross-task nearest neighbor block is in charge of mitigating the distribution difference between the source domain and the target task (given only a few examples). 

% It differs from  standard few-shot meta-learning because the test set here can leverage not only the example set, but also the original annotated set in MNLI. Standard metric-based meta learning  cannot map the new, poorly annotated classes to the existing, richly annotated classes, because the two sets are disjoint.\footnote{This is more true in NLP, because in computer vision, the new classes often share some attributes with the known classes.} 

In experiments, we apply \modelname\enspace trained on MNLI and $k$ examples from the target domain/task to two out-of-domain entailment benchmarks and two  NLP tasks (question answering and  coreference resolution). Results show the effectiveness of \modelname\enspace in addressing the challenges set forth in the two questions. Overall, we make two contributions:

\textbullet \enspace We are the first to systematically study  textual entailment in open domains, 
% and we propose an effective meta-learning solution to implement a few-shot textual entailment system. It enables an MNLI-based few-shot entailment system to perform well on other entailment benchmarks, 
given only a couple of domain-specific examples. 

\textbullet \enspace We follow the argument of some literature that textual entailment  is a unified NLP framework. Here, \textit{we make a step further by declaring that we study textual entailment not because some NLP tasks can be transformed into entailment, but because few-shot textual entailment can be a promising attempt for universal NLP when we can not guarantee the accessibility of rich annotations. }

\section{Related Work}\label{sec:relatedwork}

\paragraph{Textual Entailment.}
Textual entailment was first studied in \citet{DBLPDaganGM05} and the main focus in the early stages was to study lexical and some syntactic features. In the past few years, the research on textual entailment has been driven by the creation of large-scale datasets, such as SNLI \citep{DBLPBowmanAPM15}, science domain SciTail \citep{DBLPKhotSC18}, and multi-genre MNLI \citep{DBLPWilliamsNB18}. Representative work includes the first attentive recurrent neural network \citep{DBLPRocktaschelGHKB15} and its followers \citep{DBLPWangJ16,DBLPWangHF17}, as well as the attentive convolutional networks such as attentive pooling \citep{DBLPSantosTXZ16} and attentive convolution \citep{DBLPYinS18}, and  self-attentive large-scale language models like BERT \citep{DBLPDevlinCLT19} and RoBERTa \citep{DBLP11692}. All these studies result in systems that are overly tailored  to the datasets.

Our work differs in that we care more about few-shot applications of textual entailment, assuming that a new domain or an NLP task is not provided with rich annotated data.

\paragraph{Generalization via domain adaptation.} Two main types of domain adaptation (DA) problems  have been studied in literature: supervised DA and semi-supervised DA. In the supervised case, we have access to a large annotated  data in the source domain and a small-scale annotated data in the target domain \cite{DBLPDaume07,DBLPKangF18}. In the semi-supervised case, we have a large but \textit{unannotated} corpus in the target domain \cite{DBLPMiller19}. 

In contrast to semi-supervised DA, our work does not assume the availability of a large unlabeled data from the target domain or task. We also build more ambitious missions than the supervised DA since our work aims to adapt the model to new domains as well as new NLP tasks.

\paragraph{Generalization via few-shot learning.}
Few-shot problems are studied typically in the image domain \citep{koch2015siamese,DBLPVinyalsBLKW16,DBLPSnellSZ17,DBLPRenTRSSTLZ18,DBLPSungYZXTH18}. The core idea in metric-based few-shot learning is similar to nearest neighbors. The predicted probability of a test instance over a set of classes (i.e., only a few supporting examples are seen) is a weighted sum of classes for those supporting samples. \citet{DBLPVinyalsBLKW16} compare each test instance with those supporting examples  by the cosine distance in a method named Matching Networks. \citet{DBLPSnellSZ17} propose  Prototypical Networks which first build prototypical representations for each class by summing up representations of supporting examples, then compare classes with test instances by squared Euclidean distances. Unlike fixed metric measures, the Relation Network \citep{DBLPSungYZXTH18} implements the comparison through learning a matching metric in a multi-layer architecture.

In the language domain,  \citet{DBLPYuGYCPCTWZ18}  combine multiple metrics learned from diverse clusters of training tasks for an unseen few-shot text classification task.
\citet{DBLPHanZYWYLS18} release a few-shot relation classification dataset ``FewRel'' and compare a couple of representative methods on it. 
% This relation dataset has driven some system developments in \citep{DBLPGaoH0S19,sun9hierarchical}, and an enhanced version of the dataset ``FewRel 2.0'' \citep{gaoetal2019fewrel}. \citet{kumarcloser} study six feature space augmentation methods to improve the few-shot classification performance.

These few-shot studies assume that, in the same domain, a part of the classes have  limited samples,  while other classes  have adequate examples. In this work, we make a more challenging assumption that all classes in the target domain have only a couple of examples, and the training classes and testing classes are from different domains. 
% In addition, they often assume the disjoint of training classes and testing classes; our training classes in MNLI and testing classes from new domain/task have the same 

\paragraph{Unified natural language processing.}
% \citet{obamuyidezero} study using textual entailment to solve relation classification.
% Both \citep{DBLPKumarIOIBGZPS16}
% and
\newcite{DBLP08730} cast a group of NLP tasks as question answering over context, such as machine translation, summarization, natural language inference, sequence modeling, etc. \citet{DBLP10683} study transfer learning for broad NLP by converting every language problem into a text-to-text format. 
\citet{DBLP09286} unify question answering, text classification, and regression via span extraction to get rid of various output layers on top of BERT for different tasks. A concurrent work with ours \citep{BansalDBLP09286} studies few-shot learning in NLP, but only text classification tasks are involved.

Their unification is mainly from the perspective of system structure, i.e., some distinct NLP tasks can be converted into a common training structure. Rich annotations are still needed. 
Our entailment paradigm, instead, is driven by the fact that there is a common reasoning pattern behind \cite{DBLPDaganGM05}.
% , even though the tasks were originally believed to be very distinct from each other (e.g., coreference resolution vs. textual entailment).
In addition, we care more about the challenges in realistic scenarios where we have to handle  problems with limited annotations.

\begin{figure}[t]
\centering
\includegraphics[width=7.5cm]{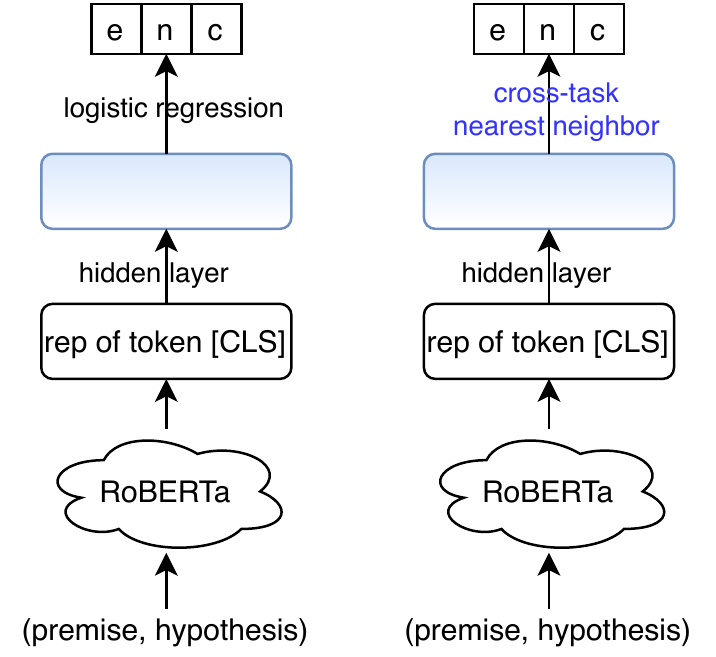}
\caption{(Left) RoBERTa for textual entailment. ``e'': entailment, ``n'': neutral, ``c'': contradiction. (Right) the skeleton of our \modelname~ system. It basically replaces the logistic regression layer in RoBERTa classifier by a cross-task nearest neighbor block. RoBERTa learns class representations implicitly in the weight matrix of logistic regression, while \modelname~ first explicitly builds class representations for both source and target tasks, then composes the cross-task probability distributions to get the prediction.} \label{fig:roberta}
\end{figure}

\section{Method}

\subsection{Problem formulation}
Provided the large-scale generic textual entailment dataset MNLI \citep{DBLPWilliamsNB18} and a few examples from a target domain or a target task,  we  build an entailment predictor that can work well in the target domain/task even if only a few examples are available.

The inputs include: MNLI, the example set (i.e., $k$ examples for each type in \{``entailment'', ``non-entailment''\} or \{``entailment'', ``neutral'', ``contradiction''\} if applicable). The output is an  entailment classifier, predicting a label for each instance in the new domain/task. Please note that we need to convert those examples into labeled entailment instances  if the target task is not a standard entailment problem. The entailment-style outputs can be easily converted to the prediction format required by the target tasks, as  introduced in Section \ref{sec:othernlp}.

\subsection{Our model \modelname}

Hereafter, we refer to MNLI as $S$ (source domain), and the new domain  or task as $T$. Before launching the introduction of  \modelname, we first give a brief description: \textit{\modelname, shown in Figure \ref{fig:roberta}, is stacking a cross-task nearest neighbor block over a \roberta~ encoder}.

\begin{figure}[t]
 \setlength{\abovecaptionskip}{0pt}
\centering
\includegraphics[width=6.5cm]{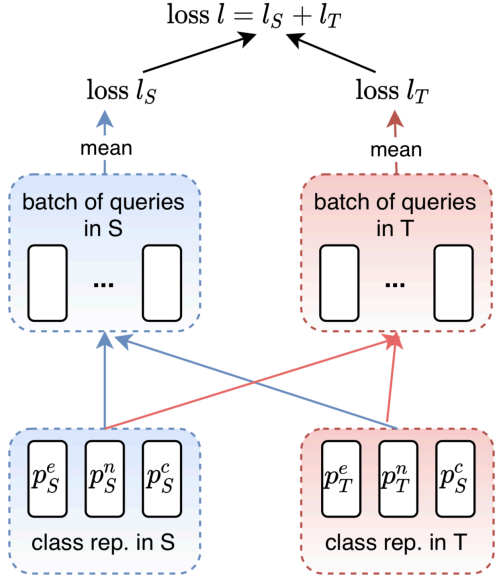}
\caption{Loss computation in cross-domain/task nearest neighbor framework.} \label{fig:loss}
\end{figure}
\begin{figure}[t]
\centering
\includegraphics[width=7.5cm]{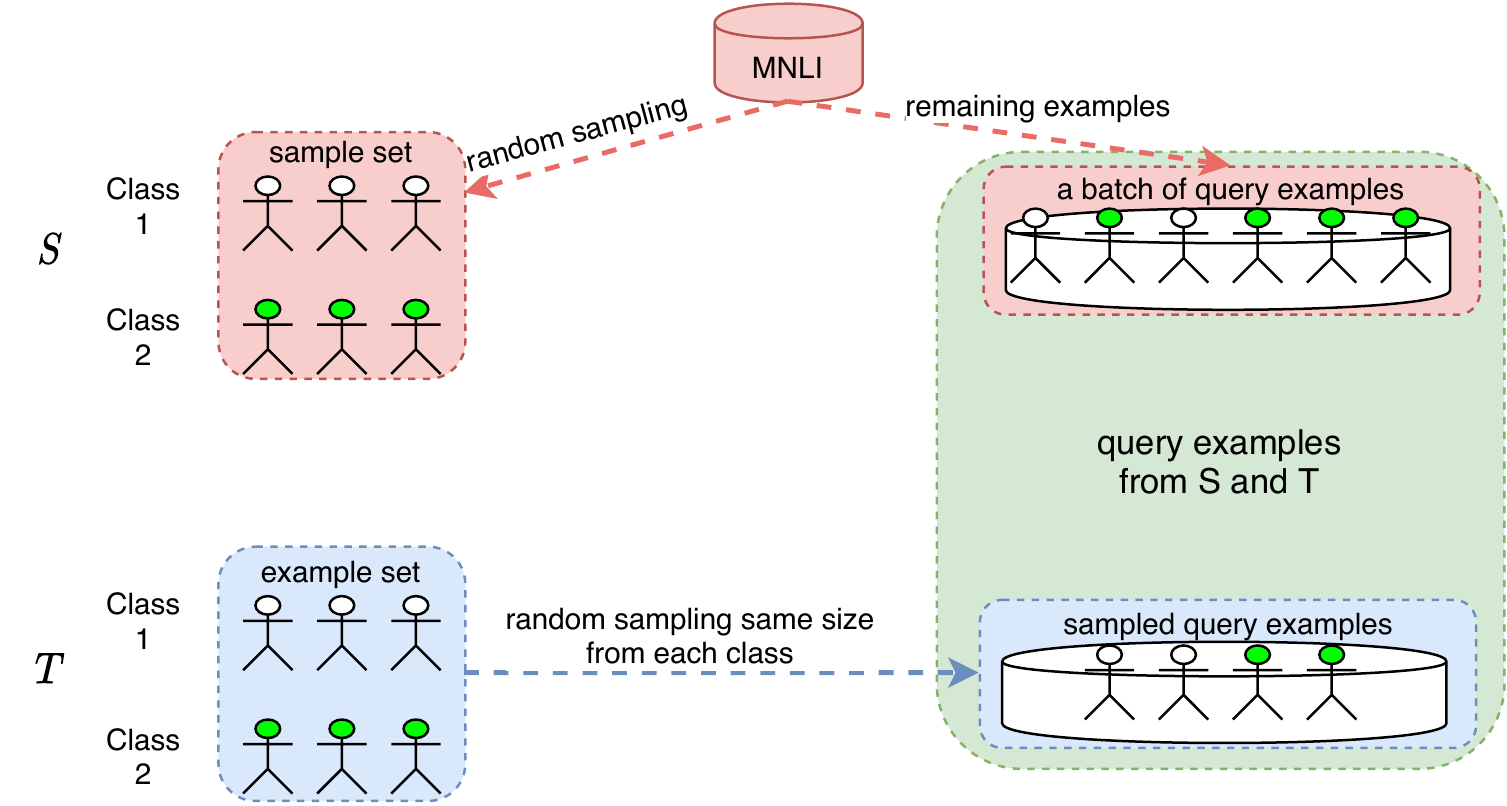}
\caption{Generating the example set in $T$, the sample set in $S$ and the query examples by both $S$ and $T$. Here we only show two classes in $S$ and $T$ just for simplicity.} \label{fig:queryexamples}
\end{figure}

\paragraph{Encoder RoBERTa.} For textual entailment, RoBERTa, shown in Figure \ref{fig:roberta}, takes the pair (premise, hypothesis) as an input. \roberta~ first outputs a representation vector (e.g., the one corresponding to the token ``CLS'') to denote the input pair, then maps this representation into a new space by a hidden layer, finally conducts  classification on that space through  logistic regression. Overall, RoBERTa works with the hidden layer together as the encoder. For convenience, we still name this ``\roberta+HiddenLayer'' encoder as ``\roberta''.

We prepare \roberta~ by  pretraining it on the source data $S$. This pretrained entailment encoder will act as a base system to deal with any new tasks (with the help of $k$ examples).

\paragraph{Cross-task nearest neighbor.} Shown in Figure \ref{fig:loss},  the first step in the cross-task nearest neighbor is to build representations for each class in the $S$ and $T$, then batches of query instances  from $S$ as well as $T$  compare with those class representations by  a matching function to compute loss and train the system. The reason we emphasize ``cross-task'' here is that both the classes and the query examples cover the two tasks $S$ and $T$. This is the core of \modelname~ in dealing with any new NLP problems of scarce annotations from textual entailment.

\textbullet\enspace\textbf{Class representations.}  We use $p_w^e$, $p_w^n$ and $p_w^c$ to denote the representations for the three classes \{``entailment'', ``neutral'', ``contradict''\} in $w$, $w \in$ \{$S$, $T$\}. When the target task $T$ can only be converted into two classes, i.e., ``entail vs. non-entail'', we let $p_T^n = p_T^c$, both denoting the class ``non-entail''.

For the target $T$, each class has $k$ labeled examples (example set). 
For the source domain $S$, similar with the episode training in meta learning \cite{DBLPSnellSZ17}, 
we  randomly sample $k$ examples (``sample set'') of each class in $S$. Then, 
\begin{equation}
    p_w^j = \frac{1}{k}\sum_{i=1}^k\mathrm{\roberta}(x_n^i)
\end{equation}
where \{$x_n^i$\}, $i=1\cdots k$, are the labeled $k$ examples for class $j\in$ \{e, n, c\}  in $T$ or   $S$,  $\mathrm{\roberta}(\cdot)\in\mathbb{R}^d$ and $p_w^j\in\mathbb{R}^d$. Overall, \modelname~ keeps representations for six classes.

\textbullet\enspace\textbf{Query examples.} As Figure \ref{fig:queryexamples} illustrates, a query batch  is composed of two sub-batches, one from  $S$, the other from $T$. For $S$, apart from its ``sample set'', the remaining labeled examples are grouped as mini-batches. For $T$, since all the labeled examples it has are those $k$ supporting examples per class, we randomly sample $m$ examples from the $k$ supporting examples for each class ($m<k$), and finally incorporate them into a $S$'s  mini-batch as a bigger batch of queries. 

Since \modelname~ is to cope with new tasks given a textual entailment task. We assume that the source entailment task provides valuable knowledge to warm up the model learning. For a testing instance in $T$, we want it to compose the reasoning conclusions derived from both $S$ and the example set in $T$.

For training, we  include examples from $S$ as queries because we treat the classes in $S$ and $T$ equally, and the queries in $S$ and $T$ equally as well. This leads to a higher-level abstract task in which  $S$ and $T$ learns from each other to mitigate the difference.

\paragraph{Matching function.} Assuming a query example gets its representation $q$ through \roberta, then a matching score, between this query example and one class (class representation $p$), $s_{p,q}$ is learnt  as follows:
\begin{align}
    I&=[p, q, p\circ q, p-q]\\
    r_1 &= \mathrm{dropout}(\mathrm{tanh}(W_1\times I))+I \\
    r_2 &= \mathrm{dropout}(\mathrm{tanh}(W_2\times r_1))+r_1\\
    r_3 &= \mathrm{dropout}(\mathrm{tanh}(W_3\times r_2))\\
    r_4 &= \mathrm{dropout}(\mathrm{tanh}(W_4\times r_3))\\
    s_{p,q} &= \mathrm{sigmoid}(W_5\times r_4)
\end{align}
where $I\in\mathbb{R}^{4d}$, $W_1$ and $W_2\in\mathbb{R}^{4d\times 4d}$, $W_3\in\mathbb{R}^{4d\times 2d}$, $W_4\in\mathbb{R}^{2d\times d}$ and  $W_5\in\mathbb{R}^{d}$. 
\begin{algorithm}[t]
\small
\SetAlgoLined
\KwIn{MNLI as $D_S$, $k\times |C|$ examples from $T$ denoted as $D_T$}
\KwOut{A 3-way entailment classifier}
Sample $k$ example from $S$'s class ``e'', ``n'' and ``c'' as $D_S^e$,  $D_S^n$ and  $D_S^c$, respectively; remaining examples in $D_S$ form minibatches $B_S$ = \{$B^i_S$\}.
%  Generate page.vocab() from sentence\_list\;

\While{each mini-batch $B^i_S$}{
build a mini-batch $B^i_T$=\{$\hat{D}_T^e$,$\hat{D}_T^n$, $\hat{D}_T^c$ \} from $D_T$, $|\hat{D}_T^i|=m$  and $m<k$.\\
build class representation: $p_w^j=\sum\mathrm{RoBERTa}(D_w^j)$, $w\in\{S, T\}$ and $j\in\{e, n, c\} $\\
build a query batch $B^i=\{B_S^i,B_T^i\}$\\
\While{each query $q$ in $B_S^i$}{
compare $q$ with class representations \{$p_w^j$\} to get probability distribution $g$\\
get loss for $q$

}
loss $l_S$ is the mean loss in $B_S^i$\\
\While{each query $q$ in $B_T^i$}{
compare $q$ with class representations \{$p_w^j$\} to get probability distribution $g$\\
get loss for $q$

}
loss $l_T$ is the mean loss in $B_T^i$\\
loss $l=l_S+l_T$ for this query batch $B^i$\\
update the \roberta~ and nearest neighbor block
}
\enspace
\caption{\modelname~ algorithm. }\label{algo:modelalgo}
\end{algorithm}
\paragraph{Probability distribution per query.} A query example will obtain three matching scores from $S$ ($g_S\in\mathbb{R}^3$) and three matching scores from $T$ ($g_T\in\mathbb{R}^3$). Now we try to combine them as a final probability distribution of thee dimensions. Instead of linear combination with artificial weights, we let the system  learn automatically the contribution of $g_S$ and $g_T$ in a new space. Therefore,  the final probability distribution $g\in\mathbb{R}^3$ is learned as follows:
\begin{align}
    \hat{g}_S &= \mathrm{sigmoid}(W_6\times g_S)\\
    \hat{g}_T &= \mathrm{sigmoid}(W_6\times g_T)\\
    \lambda &= \mathrm{sigmoid}(W_7\times [g_S,g_T])\\
    g &= \mathrm{softmax}(\lambda\circ \hat{g}_S + (1-\lambda)\circ \hat{g}_T)
\end{align}
where $W_6\in\mathbb{R}^{3}$ and $W_7\in\mathbb{R}^{6}$. $g$ is used to compute loss to train the system in training and predict the class in testing.

\paragraph{Training loss.} In training, a query batch actually contains two sub-batches, one from $S$, the other from $T$. To balance the contribution, we first compute the mean loss in $S$'s and $T$'s sub-batches respectively, obtaining $l_S$ and $l_T$, then the overall loss for that batch is $l=l_S+l_T$, demonstrated in Figure \ref{fig:loss}, 

The whole \modelname~ system is a stack of the \roberta~ and the cross-task nearest neighbor block. Its learning algorithm is summarized in the Algorithm \ref{algo:modelalgo};. \modelname~ can be trained end-to-end. 
\subsection{\modelname~ vs. other related models}

\textbullet\enspace\textbf{\modelname~ vs. Prototype. Net.} Prototypical network \cite{DBLPSnellSZ17} assumes that training tasks and test tasks are in the same distribution. So, it focuses on the matching function learning and hopes a well-trained matching function in training tasks (i.e., $S$ in this work) works well in the target tasks (i.e., $T$ here). However, the presumption does not apply to the cross-domain/task scenarios in this work. 

Similarly, \modelname~ also builds class representation by averaging the representations of some class-specific labeled examples, as prototypical network does. In training, prototypical network builds class representations in training tasks and query examples come from the training tasks only; in testing, the query examples from the testing tasks only compare with the few labeled examples specific to the testing task (training tasks do not participate anymore). In short, prototypical network only builds nearest neighbor algorithm \textit{within a task}. \modelname~ differs in that \textit{it is based on cross-task nearest neighbor} -- keeping class representations for both $S$ and $T$ in training as well as in testing; query examples in training also comes from  $S$ and $T$. Because of the mismatch of the distributions in $S$ and $T$, the goal of \modelname~ is to not only learn the matching function, but also map the instances in $S$ and $T$ to the same space.

\paragraph{\modelname~ vs. STILTS.} Given the source data $S$ and a couple of labeled examples from the target $T$, STILTS \cite{DBLP01088} first trains \roberta~ on $S$, then fine-tune on the labeled examples of $T$. Both the pretraining and fine-tuning use the same \roberta~ system in Figure \ref{fig:roberta}. It has been widely used as the state of the art technique for making use of  related tasks to improve target tasks, especially when the target tasks have limited annotations \cite{DBLP11692,DBLPSapRCBC19,DBLPClarkLCK0T19}. By the architecture, STILTS relies on the standard \roberta~ classifier which consists of a \roberta~encoder and a logistic regression on the top; \modelname~ instead has a cross-task nearest neighbor block on the top of the \roberta~encoder.

STILTS tries to learn the target-specific parameters  by tuning on the $k$ labeled examples. However, this is very challenging if $k$ is over small, like values \{1, 3, 5, 10\} we will use in our problems. We can also think STILTS learns class prototypical representations implicitly (i.e., the weights in the logistic regression layer), however, the bias term in the logistic regression layer reflect mainly the distribution in the source $S$, which is less optimal for predicting in the target $T$.

\section{Experiments}

We apply \modelname\enspace to entailment tasks of open domain and open NLP tasks. 

\paragraph{Experimental setup.} Our system is implemented with Pytorch on the transformers package released by Huggingface\footnote{\url{https://github.com/huggingface/transformers}}. We use ``RoBERTa-large'' initialized by the pretrained language model.

To mitigate the potential bias or artifacts \cite{DBLPGururanganSLSBS18} in sampling, \textit{all numbers of $k$-shot are average of five runs in  seeds \{42, 16, 32, 64, 128\}.} 

Due to  GPU memory constraints, we only update the nearest neighbor block, the hidden layer and  top-$5$ layers in \roberta. For other training configurations, please refer to our released code.

\begin{table*}[t]
\setlength{\tabcolsep}{5pt}
  \centering
  \begin{tabular}{ll|cc|cc}
     & &  \multicolumn{2}{c|}{open entailment tasks} &  \multicolumn{2}{c}{open NLP tasks} \\\hline\hline
 & &  RTE  & SciTail & QA & Coref.  \\
 \multicolumn{2}{c|}{\#entail-style pairs}  &  (2.5k) & (23k) &(4.8k) & (4k) \\\hline\hline
 & majority or random & 50.16 & 60.40 & 25.00 & 50.00\\\hline
\multirow{1}{*}{0-shot} &  train on MNLI  &  83.36 & 81.70 & 58.00  & 61.76  \\\hline

\multirow{4}{*}{1-shot} &  train on k examp. & 50.02$\pm$0.27 &  48.14$\pm$8.00 & 25.31$\pm$2.56 & 51.14$\pm$0.42  \\
&prototype network &  79.17$\pm$3.75 &  75.13$\pm$7.60 & 68.67$\pm$2.69& 61.91$\pm$17.5 \\
& STILTs  & 83.86$\pm$0.25 &  81.64$\pm$0.13 & 63.20$\pm$3.55 & 64.31$\pm$1.71 \\
& \modelname  & 84.76$\pm$0.35 &  83.73$\pm$1.10 & 71.70$\pm$2.55 & 74.20$\pm$3.14 \\\hline

\multirow{4}{*}{3-shot} &  train on k examp.  & 50.34$\pm$0.37 &  46.41$\pm$7.98 & 25.33$\pm$3.08 & 50.32$\pm$0.94 \\
&prototype network &  81.89$\pm$1.75 &  80.01$\pm$2.66 &67.90$\pm$1.53 & 63.71$\pm$21.1   \\
& STILTs & 84.02$\pm$0.54 & 81.73$\pm$0.23  &  65.28$\pm$5.60 & 64.66$\pm$2.89 \\
& \modelname &  85.06$\pm$0.34 &  83.71$\pm$1.17 &73.06$\pm$2.76 & 74.73$\pm$2.61 \\\hline

\multirow{4}{*}{5-shot} &   train on k examp.  &50.20$\pm$0.23  & 49.24$\pm$6.82 & 24.50$\pm$2.77 & 50.18$\pm$0.85 \\
&prototype network &  81.89$\pm$1.08 &  81.48$\pm$0.98 &67.50$\pm$2.34 & 73.22$\pm$0.78  \\
& STILTs & 84.15$\pm$0.47&  82.26$\pm$0.56 & 66.10$\pm$6.72& 68.25$\pm$3.49 \\
& \modelname & 84.84$\pm$0.61 & 84.82$\pm$1.18  &73.30$\pm$2.65 & 74.59$\pm$2.87 \\\hline

\multirow{4}{*}{10-shot} &   train on k examp. & 50.53$\pm$0.99 & 57.09$\pm$4.04 &  25.28$\pm$2.35 & 52.55$\pm$0.99 \\
&prototype network & 82.12$\pm$0.70  &  81.83$\pm$0.54 &68.48$\pm$2.40 & 73.28$\pm$1.51  \\
& STILTs  & 84.08$\pm$0.48 & 82.26$\pm$0.61 & 67.93$\pm$3.31 & 71.08$\pm$4.09 \\
& \modelname  & \textbf{85.28$\pm$0.27}  &  \textbf{86.19$\pm$1.10} & \textbf{74.23$\pm$2.48} & \textbf{77.58$\pm$2.50}   \\\hline\hline

\multirow{2}{*}{full-shot} &  train on target data  & 79.98$\pm$0.72 & 95.55$\pm$0.14 &80.47$\pm$3.00  & 90.20$\pm$0.45 \\
& STILTs (SOTA)  & 86.26$\pm$0.23 & 95.05$\pm$0.19  &82.60$\pm$0.64 & 89.26$\pm$0.38  \\\hline\hline
\end{tabular}
\caption{Applying \modelname\enspace to two entailment benchmarks (RTE and SciTail) and two other NLP tasks (question answering (QA) and  coreference resolution (Coref.)), each providing $k$ examples ($k=\{1,3,5,10\}$). Numbers for ``STILTS (SOTA)'' are upperbound performance while using full labeled data; bold numbers are our top numbers when the few-shot hyperparamter $k<=10$. }\label{tab:entailmentresults}
\end{table*}

\paragraph{Baselines.} The following baselines are shared by experiments on open entailment tasks and open NLP tasks.

\textbullet\enspace \textbf{0-shot.} We assume zero examples from  target domains. We train a RoBERTa classifier\footnote{Specifically, the ``RobertaForSequenceClassification'' classifier in the Huggingface transformer.} on MNLI, and apply it to  the respective test set of target domains without  fine-tuning. 

\textbullet\enspace \textbf{Train on $k$ examples.} We build a RoBERTa classifier on the k labeled examples directly. \textit{No MNLI data is used.} When $k$ is increased to cover all the labeled examples of the target domain or task, this baseline is referred as ``train on target data''.

% Using just the example set, we can in principle train a classifier to predict the label of each test instance. However, due to the limited size of the example set, the performance of such a classifier is usually unsatisfactory. 

\textbullet\enspace \textbf{STILTs \citep{DBLP01088}.} This is a learning paradigm: for any target task, first pretrain the model on intermediate tasks, then fine-tune on the target task. Here, it means pretraining on MNLI, then fine-tuning on $k$ examples ($k>= 1$ until it reaches the full labeled data of the target domain/task). When $k=0$, ``STILTS'' equals to ``0-shot'' baseline.
% STILTs shows that the supplementary training on intermediate labeled-data tasks can further improve the performance after pretraining BERT as language modeling. 

\textbullet\enspace
\textbf{Prototypical Network} \citep{DBLPSnellSZ17}. It is a representative episode-training algorithms for few-shot problems, introduced in Section \ref{sec:relatedwork}.

\textbullet\enspace\textbf{State-of-the-art.} STILTS is widely used as the state-of-the-art technique to promote the performance of a target problem with indirect supervision and task-specific fine-tuning.   According to the definition of STILTS, its paradigm is applicable to any Transformer-based models. Since RoBERTa is used as the main Transformer model, applying STILTS to RoBERTa, which pretrains on MNLI then fine-tunes on the full target data, is the state of the art for this work.

% In addition, we reimplement the popular few-shot episode-training algorithms Matching Network, Prototypical Network and Relation Network (note that all of these assume the disjoint of training classes and test classes, so we manually set the source classes and target classes in different names)

\subsection{\modelname\enspace in open domains}
We test the few-shot setting on two out-of-domain entailment datasets: GLUE RTE \citep{DBLPWangSMHLB19} and SciTail \citep{DBLPKhotSC18}. Examples in GLUE-RTE mainly come from the news and Wikipedia domains. SciTail is from the science domain, designed from the end task of  multiple-choice QA.
 Our source dataset MNLI covers a broad range of genres such as conversation, news reports, travel guides, fundraising letters, cultural articles, fiction, etc. RTE has  2,490/277/2,999 examples in train/dev/test;   SciTail has 23,596/1,304/2,126 respectively.

\subsection{\modelname\enspace in open NLP tasks}\label{sec:othernlp}
In this section, we apply \modelname\enspace as a universal framework to other distinct NLP tasks with limited annotations. An alternative approach to handle a task in which the annotations are scarce is to do transfer learning based on existing datasets of rich annotations and high relevance. However, we argue that this still results in ``training separate models for different tasks'', and it is unrealistic to presume, for $T$, that   a   related and  rich-annotation  dataset always exists. As we discussed, the final goal of NLP (or even AI) is to develop a single machine to solve diverse problems. To the end, we try few-shot  entailment here as an attempt.

For each downstream NLP task, we provide $k$ examples for helping the learning of  the textual entailment system. Next, we describe in detail how some representative NLP problems are converted to be textual entailment. Our work provides a new perspective to tackle these NLP issues, especially given only a couple of labeled examples.

\paragraph{Question Answering.}
% \begin{figure}[t]
% %  \setlength{\belowcaptionskip}{-10pt}
% %  \setlength{\abovecaptionskip}{0pt}
% \centering
% \includegraphics[width=7.5cm]{IJCAI2016_example}
% \caption{One example with 2 out of 4 questions in the
% MCTest. ``*'' marks the correct answer} \label{fig:mcexample}
% \end{figure}

% Question answering (QA) is a fundamental challenge in AI,  which  has attracted intensive studies in recently years, mainly due to the availability of large-scale labeled data and the advance of deep learning techniques. However, current neural question answering systems are overly tailored to  specific datasets---a system trained on one benchmark is often found to work poorly in another \citep{DBLPJiaL17,DBLPKhashabiKSR18,DBLPCaoRRS08}.

We attempt to handle the QA setting in which only a couple of labeled examples are provided. A QA problem can be formulated as a textual entailment problem---the document acts as the premise, and the (question, answer candidate), after converting into a natural sentence, acts as the hypothesis. Then a true (resp. false) hypothesis can be translated into a correct (resp. incorrect) answer.  We choose the QA benchmark MCTest-500 \citep{DBLPichardsonBR13} which releases an entailment-formatted corpus. MCTest-500 is a set of 500 items
(split into 300 train, 50 dev and 150 test).
Each item  consists of a document, four questions followed by
one correct answer, and three incorrect answers.

Deep learning has not achieved significant success on it because of the limited training data \cite{DBLPTrischlerYYHB16}---this is exactly our motivation that applying few-shot textual entailment to handle annotation-scarce NLP problems. 

For MCTest benchmark, we treat one question as one example. $K$-shot means we randomly sample $k$ annotated questions (each corresponds to a short article and has four answer candidates). We obtain $k$ entailment pairs for the class ``entailment'' and $3k$ pairs for the class ``non-entailment''. The official evaluation metrics in MCTest include accuracy and $\mathrm{NDCG}_4$. Here, we report accuracy.

% MCTest \citep{DBLPichardsonBR13} has two subsets. MCTest-160 is a set of 160 items (split into 70
% train, 30 dev and 60 test) and MCTest-500 a set of 500 items
% (split into 300 train, 50 dev and 150 test).
% Each item in MCTest consists of a document, four questions followed by
% one correct answer and three incorrect answers. This work reports on MCTest-500.

\paragraph{Coreference Resolution.}
Coreference resolution  aims to cluster the entities and pronouns that refer to the same object. This is a challenging task in NLP, and greatly influences the capability of machines in understanding the text.

We test on the coreference resolution benchmark GAP \citep{DBLPWebsterRAB18}, a human-labeled corpus from Wikipedia for recognizing  ambiguous pronoun-name coreference. An example from the GAP dataset is shown here:

``\textbf{McFerran}’s horse farm was named Glen View. After \underline{his} death in 1885, \textbf{John E. Green} acquired the farm.''

For a specific pronoun in the sentence, GAP provides two entity candidates for it to link. To correctly understand the meaning of this sentence, a machine must know which person (``McFerran'' or ``John E. Green'') the pronoun ``his'' refers to. GAP has such kind of annotated examples of sizes  split as 2k/454/2k in $train$/$dev$/$test$. Please note that some examples have both entity candidates as negative (201 in train, 62 in dev and 227 in testing).

In this work, we transform the coreference resolution problem   into an entailment problem by replacing the pronoun with each entity candidate. For example, the above example will lead to the following two hypotheses:

``\textbf{McFerran}’s horse farm was named Glen View. After \underline{McFerran's} death in 1885, \textbf{John E. Green} acquired the farm.'' [``entailment'']

``\textbf{McFerran}’s horse farm was named Glen View. After \underline{John E. Green's} death in 1885, \textbf{John E. Green} acquired the farm.'' [``non-entailment'']

It is worth mentioning that we append a ``'s'' to the person entity string if the pronoun is one of \{``his'', ``His'', ``her'', ``Her''\}. Otherwise, using the entity string to replace the pronoun directly.
Each replacement will yield a hypothesis---the problem ends up being predicting whether this hypothesis is correct or not, given the original sentence.

We randomly choose $k$ examples from $train$ to learn the entailment system; each example will produce two labeled entailment pairs. The GAP benchmark  evaluates the  F1 score  by gender (masculine and feminine) and the overall F1 by combining the two gender-aware F1 scores. We use the official evaluation script and report the overall F1.

\subsection{Results and Analyses}
 Table \ref{tab:entailmentresults} lists the numbers in $k$-shot settings ($k=\{1,3,5,10\}$) and the  full-shot competitor which uses the full labeled data of $T$. To start, the ``0-shot'' setting, compared with the ``majority or random'' baseline,  indicates that using MNLI as training set and test on various target $T$ has already shown some transferability; but this is far behind the SOTA. We are further interested in three main comparisons:
 
\textbullet \enspace Comparing \modelname~ with the typical metric-based meta learning approach: prototypical networks. Interestingly, prototypical network is worse than STILTS on the two entailment benchmarks while mostly outperforming STILTS slightly on QA and coreference tasks. Our system \modelname~ consistently surpasses it with big margins. Prototypical network is essentially a nearest neighbor algorithm \cite{DBLP09604} pretrained on $S$ only. A testing example in $T$ searches for its prediction by comparing with the $T$-specific class representations constructed by the $k$ examples. A pretrained nearest neighbor algorithm does not necessarily work well if $S$ and $T$ are too distinct.

\textbullet \enspace Comparing   \modelname~ with the SOTA technique STILTs in $k$-shot settings. Our algorithm   outperforms the STILTs across all the tasks. Note that  STILTs trains on $S$ and the   $k$ examples of $T$ sequentially. What STILTS does is to adapt the pretrained space to the target space, guided by $k$ examples. In contrast, \modelname~   unifies the RoBERTa encoder and the nearest neighbor algorithm by building cross-task class prototypical representations, then tries to train an unified space on  $S$ and $T$.

\textbullet \enspace Comparing   \modelname~ in $k$-shot settings with the full-shot settings. ``Full-shot'' has two systems: one pretrains on $S$ then fine-tunes on $T$, the other fine-tune on $T$ directly. Generally, we notice that pretraining on $S$ can finally promote the performance (e.g., in RTE and QA) or get similar numbers (e.g., in SciTail and Coreference tasks). Our system by 10-shot even beats the ``full-shot, train on target data'' with 5.3\% in RTE and is very close to the SOTA number by ``full-shot STILTS'' (85.28 vs. 86.26). In other three tasks (SciTail, QA, Coref.), although \modelname~ by 10-shot hasn't shown better performance than any full-shot settings, its big improvements over other 10-shot baselines across all the tasks ($\sim$4\% in SciTail, $\sim$6\% in QA and $>$4\% in coreference) demonstrate its superiority of handling open NLP problems in few-shot scenarios.

Please keep in mind that  the \modelname~system for all the reported NLP tasks  originated from the same  entailment classifier pretrained on MNLI. Our experiments  indicate: to deal with any open NLP tasks, instead of building large-scale datasets for them separately   and let  models to fit each of them, it is promising to employ a single entailment system which can generalize well with only a few annotated examples per task.

\begin{figure}[t]
\centering
\includegraphics[width=7.5cm]{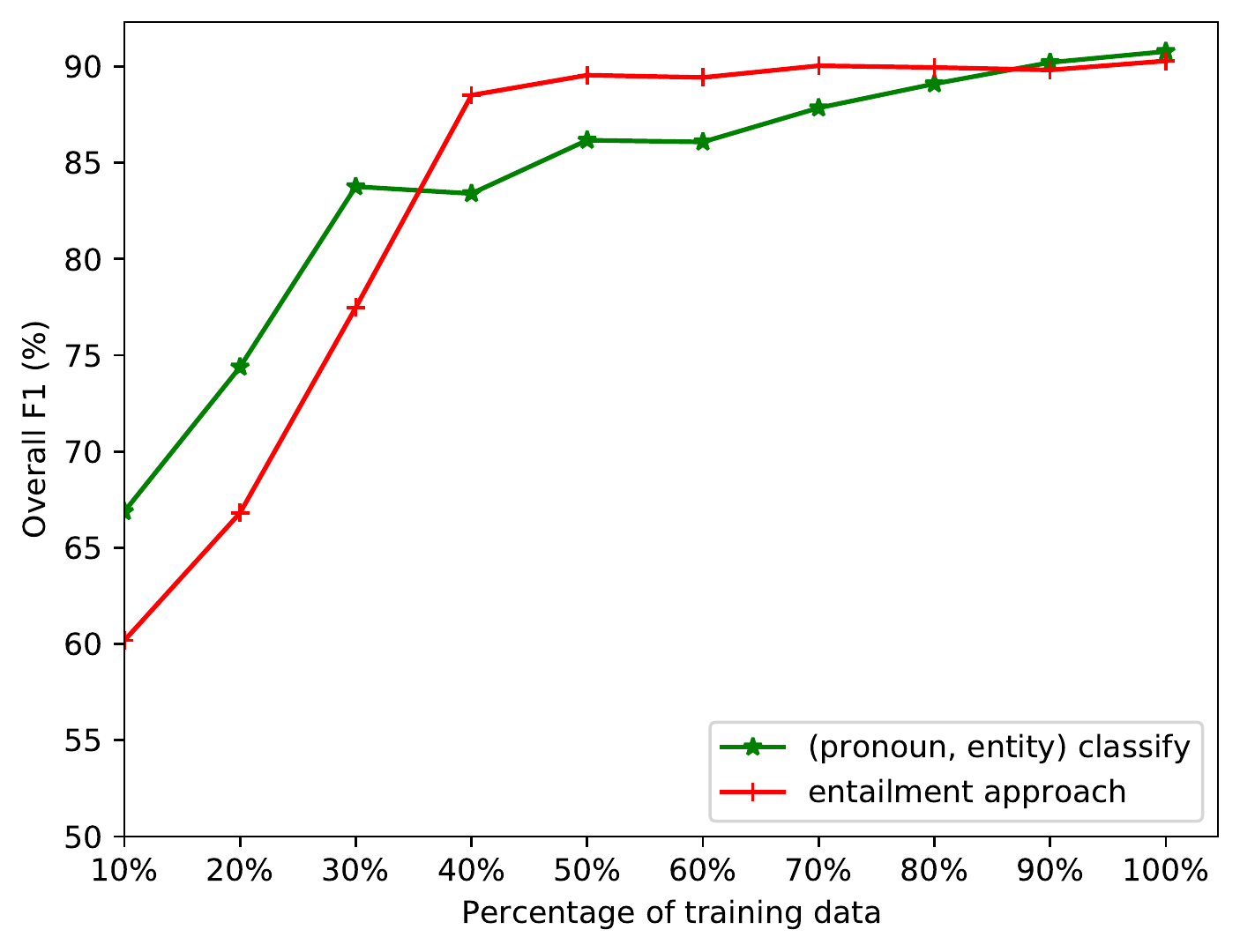}
\caption{Comparing ``entailment'' approach with non-entailment approach (i.e., classify (pronoun, entity) pairs in RoBERTa) in coreference's benchmark GAP when using different percentages of training data.} \label{fig:entailvsotheroncoref}
\end{figure}

\subsection{Reformulating  NLP problems as textual entailment: better or worse?}
In Table \ref{tab:entailmentresults}, we reported performance of dealing with open entailment and NLP tasks by entailment approach always.  We may have another question: for any NLP task, is that  better  to reformulate it as textual entailment? In this subsection, we compare textual entailment with other popular systems in modeling the coreference task which usually is not modeled in an entailment framework.

To be specific, we feed each instance in the GAP dataset into \roberta~ which will generate a representation for each token in the instance. To obtain representations for the pronoun and an entity candidate, we sum up  the representations of all tokens belonging to the pronoun or the entity string. RoBERTa is able to provide the pronoun/entity representations with context in the sentence. Finally, we do binary classification for each (pronoun, entity) pair. We compare this system with the entailment approach (i.e., ``train on target data'') when using different sizes of training set: [10\%, 20\%, $\cdots$, 100\%]. To keep a fair comparison, both systems do not pretrain on any other tasks. The result for each percentage is the average of three runs with different seeds.

Figure \ref{fig:entailvsotheroncoref} demonstrates interesting findings: (i) When using all the GAP training data, both entailment and the (pronoun, entity) classification system reach pretty similar results; (ii) When the training size is below 30\%, the non-entailment approach shows better performance. However, the entailment system converges much earlier than the competing system --- starting with 40\% training data, it can get performance almost as good as using 100\% data.

This coreference example shows that transforming an NLP task as textual entailment may obtain surprising advantages. There are more NLP tasks that can fit the entailment framework easily, such as text classification \cite{yinbenchmarking}, relation extraction, summarization, etc. However, we also need to admit that reformulating into entailment may also need to fight against new challenges. Taking text classification as an example, how to convert classification labels into hypotheses influences the results a lot. In addition, the hypothesis generation from some NLP tasks may require human efforts to guarantee the quality.

\section{Summary}
In this work, we studied how to build a textual entailment system that can work in open domains given only a couple of examples, and studied the common patterns in a variety of NLP tasks in which textual entailment can be used as a unified solver. Our goal is to push forward the research and practical use of textual entailment in a broader vision of natural language processing. To that end, we proposed utilizing MNLI, the largest entailment dataset, and a few examples from the new domain or new task to build an entailment system via  cross-task nearest neighbor. The final entailment system \modelname\enspace generalizes well to open domain entailment benchmarks and downstream NLP tasks including question answering and coreference resolution.

Our work demonstrates an example that exploring the uniform pattern behind various NLP problems, enabling us to understand the common reasoning process and create potential for  machines to learn across tasks and make easy use of indirect supervision.

\section*{Acknowledgments}

The authors would like to thank   the anonymous reviewers for insightful comments and suggestions. 

\bibliography{emnlp2020.bib}

\begin{thebibliography}{36}
\expandafter\ifx\csname natexlab\endcsname\relax\def\natexlab#1{#1}\fi

\bibitem[{Bansal et~al.(2019)Bansal, Jha, and McCallum}]{BansalDBLP09286}
Trapit Bansal, Rishikesh Jha, and Andrew McCallum. 2019.
\newblock Learning to few-shot learn across diverse natural language
  classification tasks.
\newblock \emph{CoRR}, 1911.03863.

\bibitem[{Bowman et~al.(2015)Bowman, Angeli, Potts, and
  Manning}]{DBLPBowmanAPM15}
Samuel~R. Bowman, Gabor Angeli, Christopher Potts, and Christopher~D. Manning.
  2015.
\newblock A large annotated corpus for learning natural language inference.
\newblock In \emph{{EMNLP}}, pages 632--642.

\bibitem[{Clark et~al.(2019)Clark, Lee, Chang, Kwiatkowski, Collins, and
  Toutanova}]{DBLPClarkLCK0T19}
Christopher Clark, Kenton Lee, Ming{-}Wei Chang, Tom Kwiatkowski, Michael
  Collins, and Kristina Toutanova. 2019.
\newblock Boolq: Exploring the surprising difficulty of natural yes/no
  questions.
\newblock In \emph{Proceedings of {NAACL-HLT}}, pages 2924--2936.

\bibitem[{Dagan et~al.(2005)Dagan, Glickman, and Magnini}]{DBLPDaganGM05}
Ido Dagan, Oren Glickman, and Bernardo Magnini. 2005.
\newblock The {PASCAL} recognising textual entailment challenge.
\newblock In \emph{Machine Learning Challenges, Evaluating Predictive
  Uncertainty, Visual Object Classification and Recognizing Textual Entailment,
  First {PASCAL} Machine Learning Challenges Workshop}, pages 177--190.

\bibitem[{Daum{\'{e}}~III(2007)}]{DBLPDaume07}
Hal Daum{\'{e}}~III. 2007.
\newblock Frustratingly easy domain adaptation.
\newblock In \emph{{ACL}}, pages 256--263.

\bibitem[{Devlin et~al.(2019)Devlin, Chang, Lee, and
  Toutanova}]{DBLPDevlinCLT19}
Jacob Devlin, Ming{-}Wei Chang, Kenton Lee, and Kristina Toutanova. 2019.
\newblock {BERT:} pre-training of deep bidirectional transformers for language
  understanding.
\newblock In \emph{{NAACL-HLT}}, pages 4171--4186.

\bibitem[{Gururangan et~al.(2018)Gururangan, Swayamdipta, Levy, Schwartz,
  Bowman, and Smith}]{DBLPGururanganSLSBS18}
Suchin Gururangan, Swabha Swayamdipta, Omer Levy, Roy Schwartz, Samuel~R.
  Bowman, and Noah~A. Smith. 2018.
\newblock Annotation artifacts in natural language inference data.
\newblock In \emph{{NAACL}}, pages 107--112.

\bibitem[{Han et~al.(2018)Han, Zhu, Yu, Wang, Yao, Liu, and
  Sun}]{DBLPHanZYWYLS18}
Xu~Han, Hao Zhu, Pengfei Yu, Ziyun Wang, Yuan Yao, Zhiyuan Liu, and Maosong
  Sun. 2018.
\newblock {FewRel}: {A} large-scale supervised few-shot relation classification
  dataset with state-of-the-art evaluation.
\newblock In \emph{{EMNLP}}, pages 4803--4809.

\bibitem[{Kang and Feng(2018)}]{DBLPKangF18}
Bingyi Kang and Jiashi Feng. 2018.
\newblock Transferable meta learning across domains.
\newblock In \emph{{UAI}}, pages 177--187.

\bibitem[{Keskar et~al.(2019)Keskar, McCann, Xiong, and Socher}]{DBLP09286}
Nitish~Shirish Keskar, Bryan McCann, Caiming Xiong, and Richard Socher. 2019.
\newblock Unifying question answering and text classification via span
  extraction.
\newblock \emph{CoRR}, abs/1904.09286.

\bibitem[{Khot et~al.(2018)Khot, Sabharwal, and Clark}]{DBLPKhotSC18}
Tushar Khot, Ashish Sabharwal, and Peter Clark. 2018.
\newblock {SciTaiL}: {A} textual entailment dataset from science question
  answering.
\newblock In \emph{{AAAI}}, pages 5189--5197.

\bibitem[{Koch et~al.(2015)Koch, Zemel, and Salakhutdinov}]{koch2015siamese}
Gregory Koch, Richard Zemel, and Ruslan Salakhutdinov. 2015.
\newblock Siamese neural networks for one-shot image recognition.
\newblock In \emph{ICML deep learning workshop}, volume~2.

\bibitem[{Liu et~al.(2019)Liu, Ott, Goyal, Du, Joshi, Chen, Levy, Lewis,
  Zettlemoyer, and Stoyanov}]{DBLP11692}
Yinhan Liu, Myle Ott, Naman Goyal, Jingfei Du, Mandar Joshi, Danqi Chen, Omer
  Levy, Mike Lewis, Luke Zettlemoyer, and Veselin Stoyanov. 2019.
\newblock {RoBERTa}: {A} robustly optimized {BERT} pretraining approach.
\newblock \emph{CoRR}, abs/1907.11692.

\bibitem[{McCann et~al.(2018)McCann, Keskar, Xiong, and Socher}]{DBLP08730}
Bryan McCann, Nitish~Shirish Keskar, Caiming Xiong, and Richard Socher. 2018.
\newblock The natural language decathlon: Multitask learning as question
  answering.
\newblock \emph{CoRR}, abs/1806.08730.

\bibitem[{Miller(2019)}]{DBLPMiller19}
Timothy~A. Miller. 2019.
\newblock Simplified neural unsupervised domain adaptation.
\newblock In \emph{{NAACL-HLT}}, pages 414--419.

\bibitem[{Phang et~al.(2018)Phang, F{\'{e}}vry, and Bowman}]{DBLP01088}
Jason Phang, Thibault F{\'{e}}vry, and Samuel~R. Bowman. 2018.
\newblock Sentence encoders on stilts: Supplementary training on intermediate
  labeled-data tasks.
\newblock \emph{CoRR}, abs/1811.01088.

\bibitem[{Poliak et~al.(2018)Poliak, Haldar, Rudinger, Hu, Pavlick, White, and
  Durme}]{DBLPPoliakHRHPWD18}
Adam Poliak, Aparajita Haldar, Rachel Rudinger, J.~Edward Hu, Ellie Pavlick,
  Aaron~Steven White, and Benjamin~Van Durme. 2018.
\newblock Collecting diverse natural language inference problems for sentence
  representation evaluation.
\newblock In \emph{{EMNLP}}, pages 67--81.

\bibitem[{Raffel et~al.(2019)Raffel, Shazeer, Roberts, Lee, Narang, Matena,
  Zhou, Li, and Liu}]{DBLP10683}
Colin Raffel, Noam Shazeer, Adam Roberts, Katherine Lee, Sharan Narang, Michael
  Matena, Yanqi Zhou, Wei Li, and Peter~J. Liu. 2019.
\newblock Exploring the limits of transfer learning with a unified text-to-text
  transformer.
\newblock \emph{CoRR}, abs/1910.10683.

\bibitem[{Ren et~al.(2018)Ren, Triantafillou, Ravi, Snell, Swersky, Tenenbaum,
  Larochelle, and Zemel}]{DBLPRenTRSSTLZ18}
Mengye Ren, Eleni Triantafillou, Sachin Ravi, Jake Snell, Kevin Swersky,
  Joshua~B. Tenenbaum, Hugo Larochelle, and Richard~S. Zemel. 2018.
\newblock Meta-learning for semi-supervised few-shot classification.
\newblock In \emph{{ICLR}}.

\bibitem[{Richardson et~al.(2013)Richardson, Burges, and
  Renshaw}]{DBLPichardsonBR13}
Matthew Richardson, Christopher J.~C. Burges, and Erin Renshaw. 2013.
\newblock Mctest: {A} challenge dataset for the open-domain machine
  comprehension of text.
\newblock In \emph{{EMNLP}}, pages 193--203.

\bibitem[{Rockt{\"{a}}schel et~al.(2016)Rockt{\"{a}}schel, Grefenstette,
  Hermann, Kocisk{\'{y}}, and Blunsom}]{DBLPRocktaschelGHKB15}
Tim Rockt{\"{a}}schel, Edward Grefenstette, Karl~Moritz Hermann, Tom{\'{a}}s
  Kocisk{\'{y}}, and Phil Blunsom. 2016.
\newblock Reasoning about entailment with neural attention.
\newblock In \emph{{ICLR}}.

\bibitem[{dos Santos et~al.(2016)dos Santos, Tan, Xiang, and
  Zhou}]{DBLPSantosTXZ16}
C{\'{\i}}cero~Nogueira dos Santos, Ming Tan, Bing Xiang, and Bowen Zhou. 2016.
\newblock Attentive pooling networks.
\newblock \emph{CoRR}, abs/1602.03609.

\bibitem[{Sap et~al.(2019)Sap, Rashkin, Chen, Bras, and Choi}]{DBLPSapRCBC19}
Maarten Sap, Hannah Rashkin, Derek Chen, Ronan~Le Bras, and Yejin Choi. 2019.
\newblock Social iqa: Commonsense reasoning about social interactions.
\newblock In \emph{Proceedings of {EMNLP-IJCNLP}}, pages 4462--4472.

\bibitem[{Snell et~al.(2017)Snell, Swersky, and Zemel}]{DBLPSnellSZ17}
Jake Snell, Kevin Swersky, and Richard~S. Zemel. 2017.
\newblock Prototypical networks for few-shot learning.
\newblock In \emph{{NeurIPS}}, pages 4077--4087.

\bibitem[{Sung et~al.(2018)Sung, Yang, Zhang, Xiang, Torr, and
  Hospedales}]{DBLPSungYZXTH18}
Flood Sung, Yongxin Yang, Li~Zhang, Tao Xiang, Philip H.~S. Torr, and
  Timothy~M. Hospedales. 2018.
\newblock Learning to compare: Relation network for few-shot learning.
\newblock In \emph{{CVPR}}, pages 1199--1208.

\bibitem[{Trischler et~al.(2016)Trischler, Ye, Yuan, He, and
  Bachman}]{DBLPTrischlerYYHB16}
Adam Trischler, Zheng Ye, Xingdi Yuan, Jing He, and Philip Bachman. 2016.
\newblock A parallel-hierarchical model for machine comprehension on sparse
  data.
\newblock In \emph{{ACL}}.

\bibitem[{Vinyals et~al.(2016)Vinyals, Blundell, Lillicrap, Kavukcuoglu, and
  Wierstra}]{DBLPVinyalsBLKW16}
Oriol Vinyals, Charles Blundell, Tim Lillicrap, Koray Kavukcuoglu, and Daan
  Wierstra. 2016.
\newblock Matching networks for one shot learning.
\newblock In \emph{{NeurIPS}}, pages 3630--3638.

\bibitem[{Wang et~al.(2019)Wang, Singh, Michael, Hill, Levy, and
  Bowman}]{DBLPWangSMHLB19}
Alex Wang, Amanpreet Singh, Julian Michael, Felix Hill, Omer Levy, and
  Samuel~R. Bowman. 2019.
\newblock {GLUE:} {A} multi-task benchmark and analysis platform for natural
  language understanding.
\newblock In \emph{{ICLR}}.

\bibitem[{Wang and Jiang(2016)}]{DBLPWangJ16}
Shuohang Wang and Jing Jiang. 2016.
\newblock Learning natural language inference with {LSTM}.
\newblock In \emph{{NAACL}}, pages 1442--1451.

\bibitem[{Wang et~al.(2017)Wang, Hamza, and Florian}]{DBLPWangHF17}
Zhiguo Wang, Wael Hamza, and Radu Florian. 2017.
\newblock Bilateral multi-perspective matching for natural language sentences.
\newblock In \emph{{IJCAI}}, pages 4144--4150.

\bibitem[{Webster et~al.(2018)Webster, Recasens, Axelrod, and
  Baldridge}]{DBLPWebsterRAB18}
Kellie Webster, Marta Recasens, Vera Axelrod, and Jason Baldridge. 2018.
\newblock Mind the {GAP:} {A} balanced corpus of gendered ambiguous pronouns.
\newblock \emph{{TACL}}, 6:605--617.

\bibitem[{Williams et~al.(2018)Williams, Nangia, and Bowman}]{DBLPWilliamsNB18}
Adina Williams, Nikita Nangia, and Samuel~R. Bowman. 2018.
\newblock A broad-coverage challenge corpus for sentence understanding through
  inference.
\newblock In \emph{{NAACL-HLT}}, pages 1112--1122.

\bibitem[{Yin(2020)}]{DBLP09604}
Wenpeng Yin. 2020.
\newblock Meta-learning for few-shot natural language processing: {A} survey.
\newblock \emph{CoRR}, abs/2007.09604.

\bibitem[{Yin et~al.(2019)Yin, Hay, and Roth}]{yinbenchmarking}
Wenpeng Yin, Jamaal Hay, and Dan Roth. 2019.
\newblock Benchmarking zero-shot text classification: Datasets, evaluation and
  entailment approach.
\newblock In \emph{Proceedings of the 2019 Conference on Empirical Methods in
  Natural Language Processing and the 9th International Joint Conference on
  Natural Language Processing (EMNLP-IJCNLP)}, pages 3905--3914.

\bibitem[{Yin and Sch{\"{u}}tze(2018)}]{DBLPYinS18}
Wenpeng Yin and Hinrich Sch{\"{u}}tze. 2018.
\newblock Attentive convolution: Equipping cnns with rnn-style attention
  mechanisms.
\newblock \emph{{TACL}}, 6:687--702.

\bibitem[{Yu et~al.(2018)Yu, Guo, Yi, Chang, Potdar, Cheng, Tesauro, Wang, and
  Zhou}]{DBLPYuGYCPCTWZ18}
Mo~Yu, Xiaoxiao Guo, Jinfeng Yi, Shiyu Chang, Saloni Potdar, Yu~Cheng, Gerald
  Tesauro, Haoyu Wang, and Bowen Zhou. 2018.
\newblock Diverse few-shot text classification with multiple metrics.
\newblock In \emph{{NAACL-HLT}}, pages 1206--1215.

\end{thebibliography}
\bibliographystyle{acl_natbib}

\end{document}